\documentclass[journal]{IEEEtran}
\usepackage{amsmath,amsfonts}
\usepackage{algorithm}
\usepackage{array}
\usepackage[caption=false,font=normalsize,labelfont=sf,textfont=sf]{subfig}
\usepackage{textcomp}
\usepackage{stfloats}
\usepackage{url}
\usepackage{verbatim}
\usepackage{graphicx}
\usepackage{cite}
\usepackage{algpseudocode}
\usepackage{multirow}
\usepackage{color}
\usepackage{makecell}
\usepackage{subcaption}
\usepackage{wrapfig}
\usepackage{hyperref}

\newcommand{\equ}{Eq.}
\newcommand{\tab}{Tab.}
\newcommand{\fig}{Fig.}
\newcommand{\alg}{Alg.}
\newcommand{\etal}{et al.}
\definecolor{mygray}{gray}{0.0}
\newcommand{\G}[1]{{\textcolor{mygray}{#1}}}

\begin{document}

\title{DALI: Domain Adaptive LiDAR Object Detection via Distribution-level and Instance-level Pseudo Label Denoising}

\author{Xiaohu Lu and {Hayder Radha,~\IEEEmembership{Fellow,~IEEE} }
\thanks{Michigan State University, 
East Lansing, MI 48824, United States
{\tt\small (luxiaohu,radha)@msu.edu}}
}



\maketitle

\begin{abstract}
Object detection using LiDAR point clouds relies on a large amount of human-annotated samples when training the underlying detectors' deep neural networks. However, generating 3D bounding box annotation for a large-scale dataset could be costly and time-consuming. Alternatively, unsupervised domain adaptation (UDA) enables a given object detector to operate on a novel new data, with unlabeled training dataset, by transferring the knowledge learned from training labeled \textit{source domain} data to the new unlabeled \textit{target domain}. Pseudo label strategies, which involve training the 3D object detector using target-domain predicted bounding boxes from a pre-trained model, are commonly used in UDA. However, these pseudo labels often introduce noise, impacting performance. In this paper, we introduce the Domain Adaptive LIdar (DALI) object detection framework to address noise at both distribution and instance levels. Firstly, a post-training size normalization (PTSN) strategy is developed to mitigate bias in pseudo label size distribution by identifying an unbiased scale after network training. To address instance-level noise between pseudo labels and corresponding point clouds, two pseudo point clouds generation (PPCG) strategies, ray-constrained and constraint-free, are developed to generate pseudo point clouds for each instance, ensuring the consistency between pseudo labels and pseudo points during training. We demonstrate the effectiveness of our method on the publicly available and popular datasets KITTI, Waymo, and nuScenes. We show that the proposed DALI framework achieves state-of-the-art results and outperforms leading approaches on most of the domain adaptation tasks. Our code is available at \href{https://github.com/xiaohulugo/T-RO2024-DALI}{https://github.com/xiaohulugo/T-RO2024-DALI}.

\end{abstract}


\section{Introduction}
3D object detection using LiDAR point clouds for autonomous driving and related applications ~\cite{wu2022sparse,deng2021voxel,shi2021pv,shi2020point,pang2020clocs} is highly dependent upon the availability of a large amount of human-annotated training samples. However, generating annotations of 3D bounding boxes for object detection is usually a tedious, quite inefficient, and costly process. The lack of easy-to-use annotation tools aggravates the problem further, especially for 3D point clouds labeling. Meanwhile, the generalization ability of deep neural networks trained on a specific \textit{source} dataset is poor when applied to different \textit{target} domains. Therefore, developing a strategy that can take advantage of existing annotated datasets to facilitate 3D object detection on unannotated datasets has significant benefits in both academia and industry.
    \begin{figure}
    	\centering
    	\footnotesize
    	\begin{tabular}{c}
    		\includegraphics[width=0.90\linewidth]{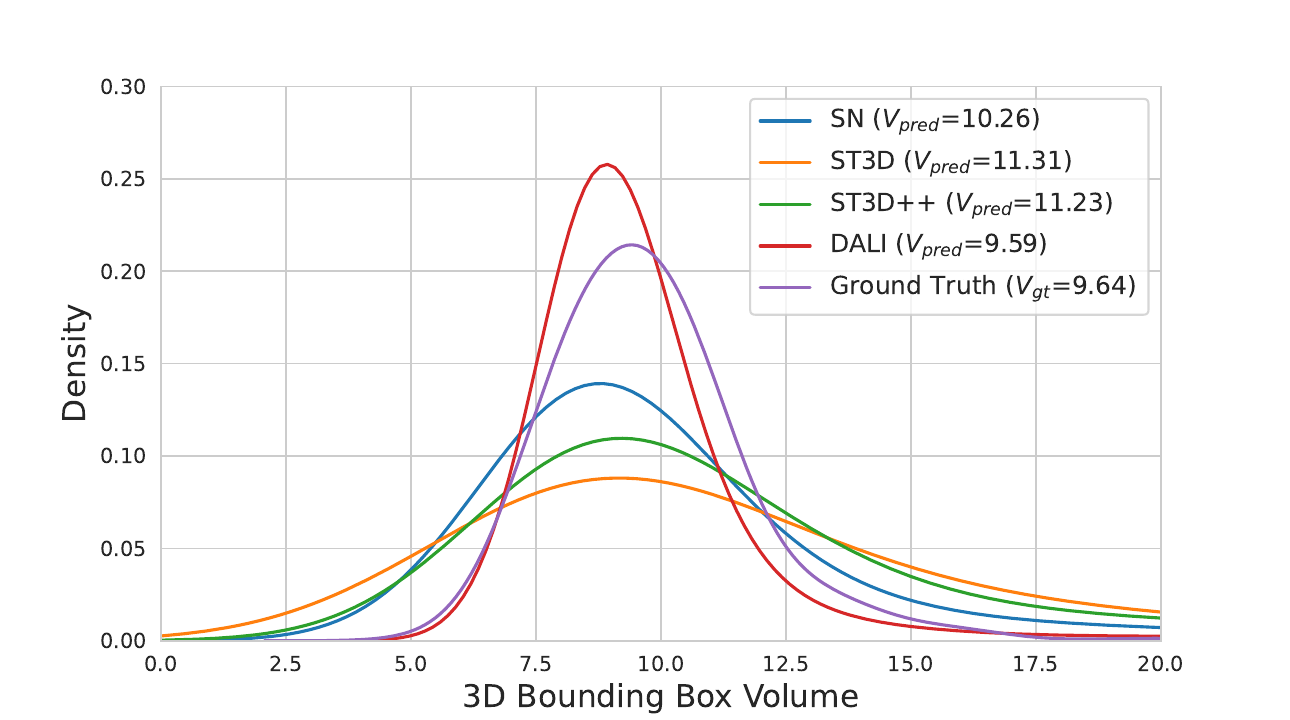} \\
                (a) Distribution-level noise\\
    		\includegraphics[width=0.90\linewidth]{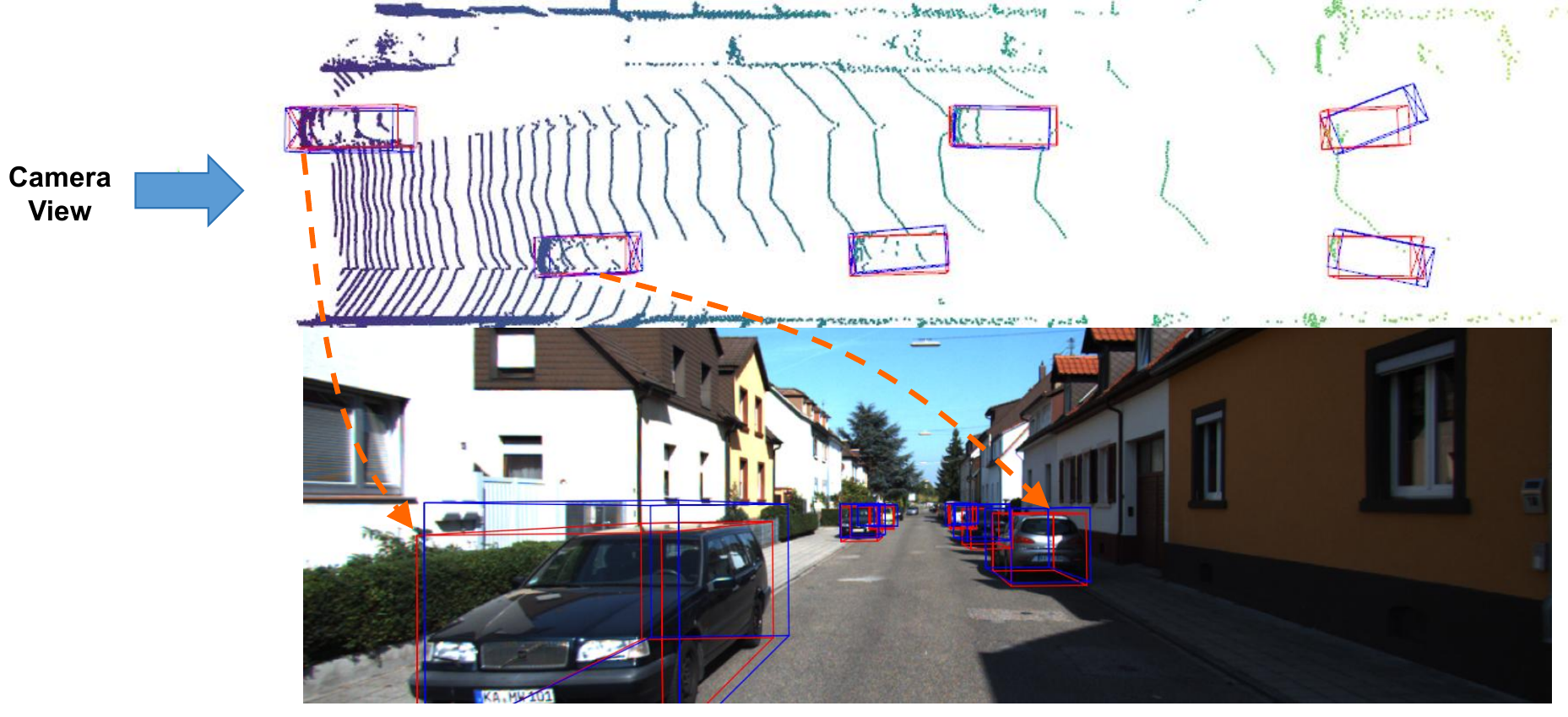} \\
                (b) Instance-level noise\\
    	\end{tabular} 
    	\caption{(a) illustrates the distributions of pseudo label volumes across various methods, alongside the ground truth distribution. In (b), an instance-level noise example of pseudo labels is presented. Both (a) and (b) are generated within the context of the nuScenes→KITTI domain adaptation task, utilizing SECOND-IoU as the backbone. Notably, our method, DALI, exhibits the smallest divergence from the ground truth distribution.}
    	\label{fig:distribution}
    \end{figure} 
    
Unsupervised domain adaptation (UDA)~\cite{zhao2020review,wilson2020survey} tackles this problem by training the deep neural network on the annotated source domain and then adapting the learned knowledge to the target domain by reducing the divergence between the two domains' distributions. Typical methods in UDA include: 1) adversarial domain adaptation~\cite{ganin2016domain,tzeng2017adversarial} that introduces a domain classifier to align the source and target domains by adversarial training, 2) distribution discrepancy minimization~\cite{tzeng2014deep,long2015learning} that minimizes the discrepancy between the source and target domain distributions, 3) disentangled representation learning~\cite{li2021unsupervised,lee2021dranet} that learns a shared feature representation space while disentangling domain-specific factors, and 4) pseudo labeling~\cite{gu2020spherical,ge2020mutual} that use the predicted pseudo labels generated from the pre-trained model to train a target-specific network. Among these methods, pseudo labeling is widely used in both 2D~\cite{yu2022sc,li2021category} and 3D~\cite{yang2021st3d,yang2022st3d++} object detection because of its simplicity of implementation. However, pseudo labels are usually noisy due to the lack of ground truth annotations. Some of the previous methods try to address this problem by refining the pseudo labels, for example in ST3D~\cite{yang2021st3d} and ST3D++~\cite{yang2022st3d++}, historical pseudo labels are assembled via a strategy called hybrid quality-aware triplet memory (HQTM) to improve the quality and stability of the pseudo labels. However, pseudo label denoising remains an open problem in domain adaptive object detection.
    
In the scenario of LiDAR object detection, the noise in pseudo labels usually comes from two sources: the distribution-level noise and the instance-level noise. The distribution-level noise exists when the distribution of the pseudo labels, namely 3D bounding boxes, is deviated from the actual distribution of the target dataset. \fig~\ref{fig:distribution}(a) shows the volume distribution curves of pseudo labels generated by different methods and the ground truth one on the KITTI dataset in nuScenes→KITTI domain adaptation task. We can see that even though the size normalization strategies like random object scaling~\cite{yang2021st3d} and statistic normalization~\cite{wang2020train} are adopted by previous methods like SN~\cite{wang2020train}, ST3D~\cite{yang2021st3d}, ST3D++~\cite{yang2022st3d++}, the actually predicted mean object sizes are significantly deviated from the ground truth, which highlights the importance of finding an effective method to address the distribution-level noise of pseudo labels. Instance-level noise, illustrated in \fig~\ref{fig:distribution}(b), occurs when the pseudo label does not align with the corresponding point clouds. This inconsistency is mainly attributed to the sparsity of point clouds and the inadequacy of training samples.

In this paper, we propose new strategies to address the pseudo label denoising problem that is due to both distribution-level and instance-level sources. 
To mitigate distribution-level noise, we introduce a post-training size normalization (PTSN) strategy. This approach aims to align the predicted mean object size with the ground truth mean object size by identifying the optimal unbiased scale that minimizes the difference between predicted and ground truth mean object sizes. To reduce the instance-level noise, we collect a library of 3D models and a library of LiDAR sensors, and develop two types of pseudo point clouds generation (PPCG) strategies, ray-constrained and constraint-free, to generate noise-free pseudo point clouds by simulating the process of LiDAR scanning the 3D models. Finally, the pseudoly labeled target domain and the labeled source domain are used to train a unified domain adaptive network capable of performing effectively in both the source and target domains. Experimental results on popular 3D object detection datasets like KITTI~\cite{geiger2012we}, Waymo~\cite{sun2020scalability}, and nuScenes~\cite{caesar2020nuScenes} demonstrate the effectiveness of our approach. In summary, our contributions are as follows:
    \begin{itemize}
      \item We propose a novel post-training size normalization (PTSN) strategy to address the distribution-level noise of pseudo labels by identifying an optimal unbiased scale after network training.
      \item We develop two types of pseudo point clouds generation (PPCG) strategies, ray-constrained and constraint-free, which generate new virtually noise-free pseudo point cloud samples for the object under consideration to address the instance-level pseudo label noise.
      \item Unlike state-of-the-art methods like ST3D~\cite{yang2021st3d} and ST3D++~\cite{yang2022st3d++} that only works well on the target domain, our method can achieve competitive results in both the source and target domains.
      \item Experimental results on three popular 3D object detection datasets, i.e.,  KITTI~\cite{geiger2012we}, Waymo~\cite{sun2020scalability}, and nuScenes~\cite{caesar2020nuScenes}, show that our method achieves better performance than state-of-the-art methods. 
    \end{itemize}

\section{Related Work}
    \begin{figure*}
    	\centering
    	\footnotesize
    	\begin{tabular}{c}
    		\includegraphics[width=0.97\linewidth]{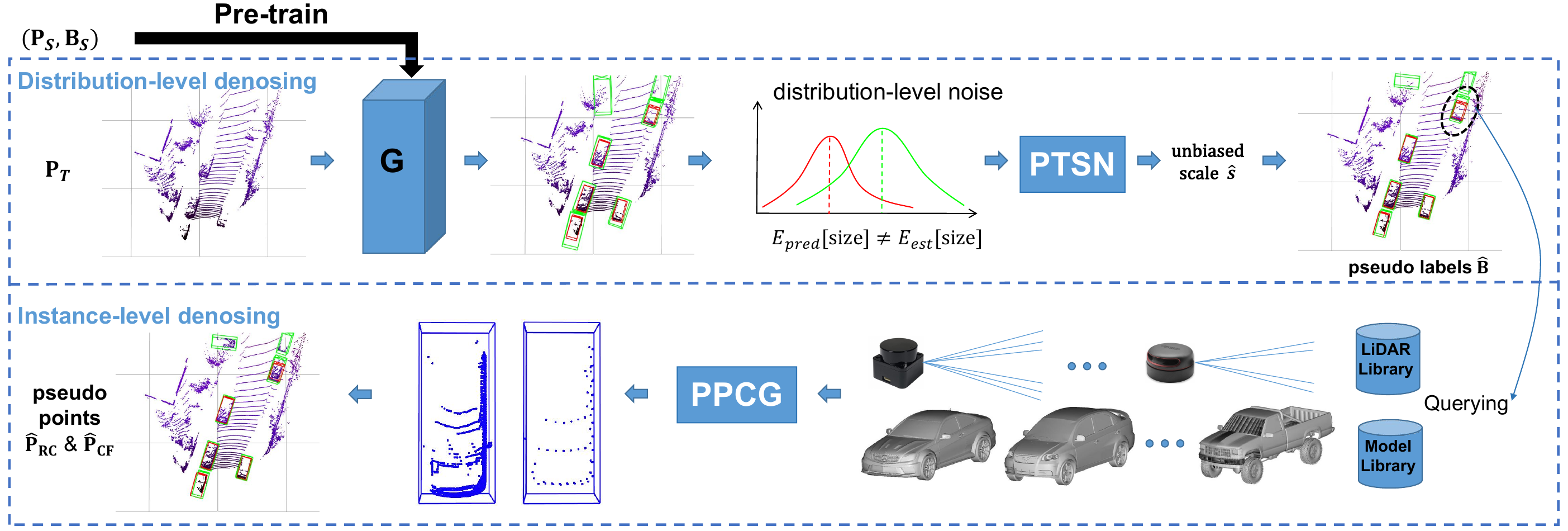}
    	\end{tabular} 
    	\caption{The framework of our domain adaptive LiDAR object detection method. Given the network \textbf{G} trained on the labeled source samples $(\textbf{P}_\text{S}, \textbf{B}_\text{S})$ and the estimated ground truth mean object size $E_{est}[\text{Size}]$ obtained from SN~\cite{wang2020train} or ROS~\cite{yang2021st3d}, we develop post-training size normalization (PTSN) to address the distribution-level noise by selecting the optimal unbiased scale that makes $E_{pred}[\text{Size}] \approx E_{est}[\text{Size}]$. Then the pseudo bounding boxes $\hat{\textbf{B}}$ are generated accordingly, and a pseudo point clouds generation (PPCG) strategy is proposed to address the instance-level noise of $\hat{\textbf{B}}$ by generating two types of pseudo points $\hat{\textbf{P}}_{\text{RC}}$ and $\hat{\textbf{P}}_{\text{CF}}$ for $\hat{\textbf{B}}$ based on a 3D model library and a LiDAR sensor library. Finally, $(\textbf{P}_\text{S}, \textbf{B}_\text{S})$, 
    	$(\hat{\textbf{P}}_{\text{RC}}, \hat{\textbf{B}})$, and $(\hat{\textbf{P}}_{\text{CF}}, \hat{\textbf{B}})$ are utilized to train a new network which is able to perform well in both source domain and target domain.}
    	\label{fig:framework}
    \end{figure*} 
    
\textbf{3D Object Detection.}
3D object detection is a rapidly evolving field in computer vision, and researchers have proposed numerous techniques to extract meaningful information from 3D data. Voxel-based methods, which represent the scene as a 3D voxel grid, have been widely used in recent years due to their ability to handle large-scale scenes and extract 3D features efficiently. SECOND~\cite{yan2018second} is one of the pioneering works in this category, where the authors proposed to divide the scene into 3D voxels and apply fast 3D sparse convolution to extract features. DSVT~\cite{wang2023dsvt}, VoxelNet~\cite{zhou2018voxelnet}, VoxelRCNN~\cite{deng2021voxel}, SE-SSD~\cite{zheng2021se} are other notable works that employ similar techniques. 
On the other hand, point-based methods, which directly operate on raw point clouds, have also shown remarkable performance in 3D object detection. PointRCNN~\cite{shi2019pointrcnn}, for instance, proposes a two-stage framework that first extracts 3D features from point clouds and then predicts object bounding boxes. 3DSSD~\cite{yang20203dssd}, PointGNN~\cite{shi2020point} are other popular point-based methods that have shown competitive performance.
In recent years, voxel-point-based methods have emerged as a promising direction, which combines the advantages of both voxel-based and point-based methods to improve detection accuracy and efficiency. PV-RCNN++~\cite{shi2023pv} and PV-RCNN~\cite{shi2020pv} are two representative methods that adopt 3D sparse convolution in voxel-based methods and point set abstraction in point-based methods to extract features. SA-SSD~\cite{he2020structure} is another notable work in this category that designs
an auxiliary network to convert the convolutional voxel features in the backbone network back to point-level representations.

\textbf{Unsupervised Domain Adaptation.}
The primary objective of unsupervised domain adaptation is to transfer the knowledge learned from the labeled source domain to the unlabeled target domain by reducing the domain divergence. One method to achieve this goal is distribution alignment, which attempts to minimize the distance between distributions in the source domain and target domain. For example, Tzeng~\etal~\cite{tzeng2014deep} used one kernel function and one adaptation layer for moment matching, Long~\etal~\cite{long2015learning} extended this strategy to multi kernels and multi adaptation layers leading to better performance. Similarly, in the work of Peng~\etal~\cite{peng2019moment}, this idea was further applied to the multi-source domain adaptation. Feature alignment methods~\cite{hsu2020progressive,chen2019progressive} are also popular strategies in this branch. Another popular paradigm in unsupervised domain adaptation is to leverage the idea of adversarial learning to domain adaptation by training a binary domain classifier to distinguish the data points from different domains. Based on this idea, most of the following methods~\cite{chen2022reusing,xia2021adaptive,ganin2016domain,zhao2024deep,awais2021adversarial,gao2021gradient,huang2021rda} explored additional information to further reduce domain divergence. For example, in the work of Long~\etal~\cite{long2018conditional}, the prediction scores were combined with the feature vector of the data as conditional information to align the domains. While in the work of Chen~\etal~\cite{chen2019transferability}, the eigenvectors with larger singular values were employed to enhance the transferability of the network. Other methods like pseudo label based methods~\cite{hu2021simple,lee2013pseudo}, regularization and normalization based methods~\cite{abuduweili2021adaptive,cai2021exponential}, and data augmentation based methods~\cite{na2021fixbi} are also popular research approaches in unsupervised domain adaptation.

\textbf{Domain Adaptive Object Detection.}
Recently, unsupervised domain adaptation on 2D object detection has been extensively studied~\cite{chen2018domain,he2022cross,wu2022target,zhao2022task,zhou2022multi,li2022sigma,hnewa2023integrated}. Some methods employ the well-developed unsupervised domain adaptation strategies in image classification to object detection. For example, Chen~\etal~\cite{chen2018fasterrcnn} combine the adversarial learning strategy~\cite{xia2021adaptive,ganin2016domain} with the classical Faster R-CNN~\cite{ren2015faster}, while in the work of Hnewa and Radha~\cite{hnewa2023integrated} and the work of Zhou~\etal~\cite{zhou2023ssda}, another popular 2D detector YOLO is utilized. Other methods like TDD~\cite{he2022cross} proposes a novel target-perceived dual-branch distillation strategy to integrate detection branches of both source and target domains in a unified teacher-student learning scheme to reduce domain shift in 2D object detection. And 
IRG~\cite{vibashan2023instance} proposes to train an instance relation graph network to guide the contrastive representation learning for domain adaptation.

Meanwhile, unsupervised domain adaptation for 3D object detection is also widely explored in the literature~\cite{wang2020train,saltori2020sf,yang2021st3d,xu2021spg,luo2021unsupervised,you2022exploiting,yang2022st3d++,tsai2022see,wei2022lidar,peng2023cl3d,hu2023density,li2023adaptation}. In the early work of Wang~\etal~\cite{wang2020train}, a strategy known as statistical normalization is developed to resize the source domain bounding box annotations to approximate those in the target domain. Similarly, in both ST3D~\cite{yang2021st3d} and ST3D++~\cite{yang2022st3d++}, a random object scaling strategy along with a hybrid quality-aware triplet memory based pseudo label denoising method are proposed to address the domain adaptation problem in 3D object detection. Moreover, Xu~\etal~\cite{xu2021spg} proposes to generate semantic points at the predicted foreground regions to recover missing points, which is then used by LiDAR-based detectors to improve the detection results. Luo~\etal~\cite{luo2021unsupervised} develop a multi-level consistency network to explore the point-, instance- and neural statistics-level consistency within a teacher-student framework. You~\etal~\cite{you2022exploiting} take advantage of nearby objects' tracks and smooth them forward and backward
in time to interpolate and extrapolate detections for missing objects due to occlusions or far ranges. Tsai~\etal~\cite{tsai2022see} generate triangle meshes based on the foreground point clouds first and then resample point clouds from these meshes with the same pattern in both source and target domains to reduce domain divergence. Wei~\etal~\cite{wei2022lidar} develop a novel LiDAR distillation strategy to generate low-beam pseudo LiDAR progressively by down-sampling the high-beam point clouds. More recently, Peng~\etal~\cite{peng2023cl3d} propose to use spatial geometry alignment and temporal motion alignment to generate feature-fusion prototype representation to align the source domain and target domain. Hu~\etal~\cite{hu2023density} develop a density-insensitive domain adaption framework to address the density-induced domain gap.

\section{Methodology}


\subsection{Distribution-level Denoising}
\label{distribution_level_denoising}
Statistical normalization (SN)~\cite{wang2020train} and random object scaling (ROS)~\cite{yang2021st3d} are two methods used to address the divergence of annotation's size distribution between two datasets. They modify the annotations of 3D bounding boxes in the source domain to make them more similar to those in the target domain. Consequentially, a network trained on the modified source domain will work better on the target domain. In SN, the deviation between the mean object sizes, namely the length, width, and height, in the source and target domains is calculated and added to the bounding box annotation of each object in the source domain. Similarly, ROS multiplies a uniformly distributed scale to the source bounding box annotation to make the mean object size in the source domain approximate to that in the target domain. However, the modification on the source domain cannot guarantee that the predicted mean object size in the target domain will still be equal to the ground truth mean object size. As shown in \tab~\ref{tab:sn_ros_ptsn}, there exist significant gaps between the predicted mean object size and ground truth mean object size for both SN and ROS, indicating that neither method can achieve the expected performance on reducing the distribution-level noise of the pseudo labels for the target domain.

    \begin{table*}[!ht]
        \centering
    	\caption{The ground truth mean object size $E_{gt}[\text{Size}]$, the estimated ground truth mean object size $E_{est}[\text{Size}]$, and the predicted mean object size $E_{pred}[\text{Size}]$ of the target domain (KITTI) pseudo labels of different methods: ROS~\cite{yang2021st3d}, SN~\cite{wang2020train}, and our PTSN on Waymo → KITTI domain adaptation task with SECOND-IoU \cite{yang2021st3d} as the backbone.}
    	\label{tab:sn_ros_ptsn}
    	\setlength{\tabcolsep}{3.0mm}
    	\begin{tabular}{l c c c c}
    	\hline
    	Method &$E_{gt}[\text{Size}]$ &$E_{est}[\text{Size}]$ &$E_{pred}[\text{Size}]$ &Size Bias \\
    	\hline
            ROS~\cite{yang2021st3d} &[3.89, 1.62, 1.53] &[3.88, 1.73, 1.45]  &[4.19, 1.86, 1.60]  &[0.30, 0.24, 0.07]  \\
            SN~\cite{wang2020train} &[3.89, 1.62, 1.53] &[3.89, 1.62, 1.53]  &[4.11, 1.71, 1.61]  &[0.21, 0.09, 0.08]  \\
            PTSN(w/ ROS)            &[3.89, 1.62, 1.53] &[3.88, 1.73, 1.45]  &[3.87, 1.72, 1.50]  &[-0.02, 0.10, -0.03] \\
            PTSN(w/ SN)             &[3.89, 1.62, 1.53] &[3.89, 1.62, 1.53]  &[3.84, 1.70, 1.50]  &[-0.05, 0.08, -0.03]  \\
    	\hline
        \end{tabular}
    \end{table*}
    
    \begin{figure}
    	\centering
    	\footnotesize
    	\begin{tabular}{c}
    		\includegraphics[width=0.95\linewidth]{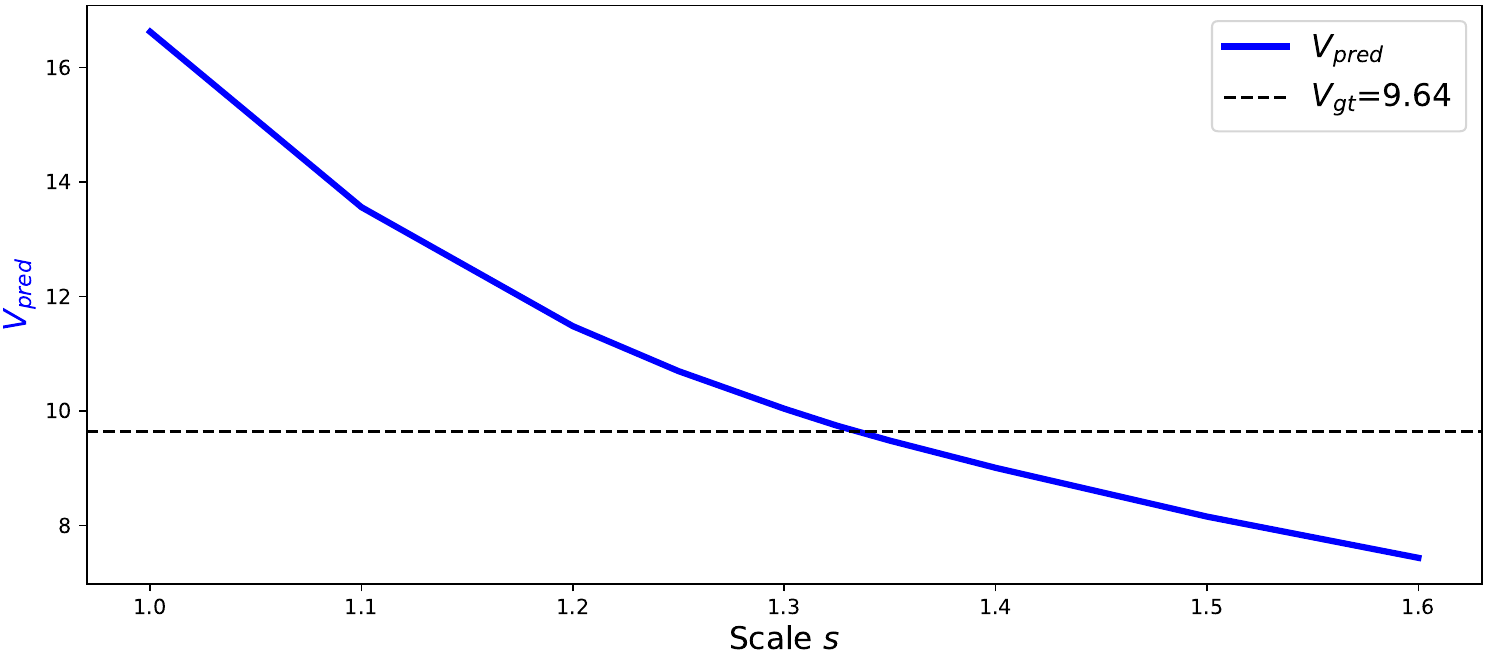}    	\end{tabular} 
    	\caption{Volume of $E_{pred}[\text{Size}](s)$ of the car class on KITTI \textit{val} of different scales given the SECOND-IoU \cite{yang2021st3d} network pre-trained on Waymo dataset.}
    	\label{fig:psn}
    \end{figure} 
    
\subsubsection{Post-training Size Normalization via Scaling}
As we can learn from both SN and ROS, the difference in object size between the source and target domains is the major factor that leads to the distribution-level noise of the pseudo labels. Therefore, we propose a novel post-training size normalization (PTSN) strategy that focuses on finding the optimal unbiased scale for the target domain such that the predicted mean object size is as close as possible to the ground truth one. More specifically, given the network \textbf{G} pre-trained on the source domain, for the $i_{th}$ frame in the target domain, we first multiply the X, Y, and Z coordinates of its point clouds $\text{p}_T^i$ by a scale $s$, and then forward the scaled point clouds to \textbf{G} to get the pseudo bounding boxes. Finally, those pseudo bounding boxes are scaled back to the original coordinate system by dividing the location coordinates (x, y, z) and dimensions (l, w, h) of the pseudo bounding boxes by the same scale $s$. Note that the direction of the pseudo bounding box is invariant in response to the changing scale. Therefore, we can use the original value of the predicted direction directly. As a summary, given the pre-trained network \textbf{G}, the predicted mean object size at the scale $s$ can be calculated as below:
    \begin{equation}\label{equ:psn}
    E_{pred}[\text{Size}](s) = \frac{1}{N_t} \sum_{i=1}^{N_t} \frac{1}{O_i} \sum_{j=1}^{O_i}  \frac{1}{s} \text{Size} \left( \textbf{G}_j(s \times \text{p}_T^i) \right),
    \end{equation}
where $N_t$ is the number of frames in the target domain, $O_i$ is the number of objects detected in the $i^{th}$ target-domain frame, and $\frac{1}{s} \text{Size} \left( \textbf{G}_j(s \times \text{p}_T^i) \right)$ denotes the size of the $j^{th}$ pseudo bounding box scaled back to the original target domain dimensions.

In \fig~\ref{fig:psn}, we observe the relationship between the scale factor $s$ and the volume of the predicted mean object size on the KITTI dataset, given the SECOND-IoU \cite{yang2021st3d} network pre-trained on the Waymo dataset. The figure shows that the average value of the predicted volume $V_{pred}$ decreases as the scale factor increases because the pre-trained network \textbf{G} generates bounding boxes $\textbf{G}(s \times \text{p}_T^i)$ according to the object's distribution in the source domain, which causes $E_{pred}[\text{Size}](s)$ to be approximated by $1/s$ of the mean object size in the source domain. Given the monotone characteristic of \equ~\ref{equ:psn}, our post-training size normalization method tries to find out the optimal value of scale $\hat{s}$ from a list of scale hypotheses to make the predicted mean object size approximate the ground truth mean in the target domain. However, in domain adaptive LiDAR object detection, the ground truth mean target object size is unknowable. Fortunately, both SN~\cite{wang2020train} and ROS~\cite{yang2021st3d} provide us solutions to estimate the ground truth mean size. As shown in \tab~\ref{tab:sn_ros_ptsn}, the estimated ground truth mean size $E_{est}[\text{Size}]$ is close to the ground truth mean size $E_{gt}[\text{Size}]$. As a summary, given the estimated ground truth mean size $E_{est}[\text{Size}]$ obtained from SN or ROS, our proposed post-training size normalization (PTSN) tries to find out the optimal unbiased scale $\hat{s}$ among a set of scales to make the predicted mean object size equals to $E_{est}[\text{Size}]$.

\subsection{Instance-level Denoising}

Even though we can find out the optimal unbiased scale $\hat{s}$ to reduce the distribution-level noise, some of the pseudo labels can still be noisy due to the sparsity of point clouds. 
Unlike ST3D~\cite{yang2021st3d} and ST3D++~\cite{yang2022st3d++} that endeavor to refine the pseudo labels for the point clouds, we propose to generate \textit{pseudo point clouds} for the pseudo bounding boxes such that both the points and bounding boxes can be consistent (noiseless) with each other for training.

To achieve this goal, we adopt a virtual simulation procedure that utilizes a virtual LiDAR sensor to scan 3D models. We start by collecting the configuration information of different LiDAR sensors used in Waymo, KITTI, and nuScenes to form a library $\mathcal{L}$, as shown in \tab~\ref{tab:datasets_overview}. To handle different cases for 3D models, we propose two strategies: \textit{CAD-based} and \textit{point-based}. When detailed CAD models are available, such as in the CADillac car dataset~\cite{cadillac}, we select high-quality models to form a 3D model library $\mathcal{M}$. Otherwise, we collect strong instances of objects that have dense internal points (details are introduced in Sec.~\ref{ExperimentalSetup}) from the annotated source domain to form the 3D model library $\mathcal{M}$. Given the model library $\mathcal{M}$ and sensor library $\mathcal{L}$, we develop two types of pseudo point clouds generation (PPCG) strategies: ray-constrained and constraint-free to generate noise-free pseudo point clouds to address the instance-level pseudo label noise.

    \begin{figure*}
    	\centering
    	\footnotesize
    	\begin{tabular}{c}
    		\includegraphics[width=0.95\linewidth]{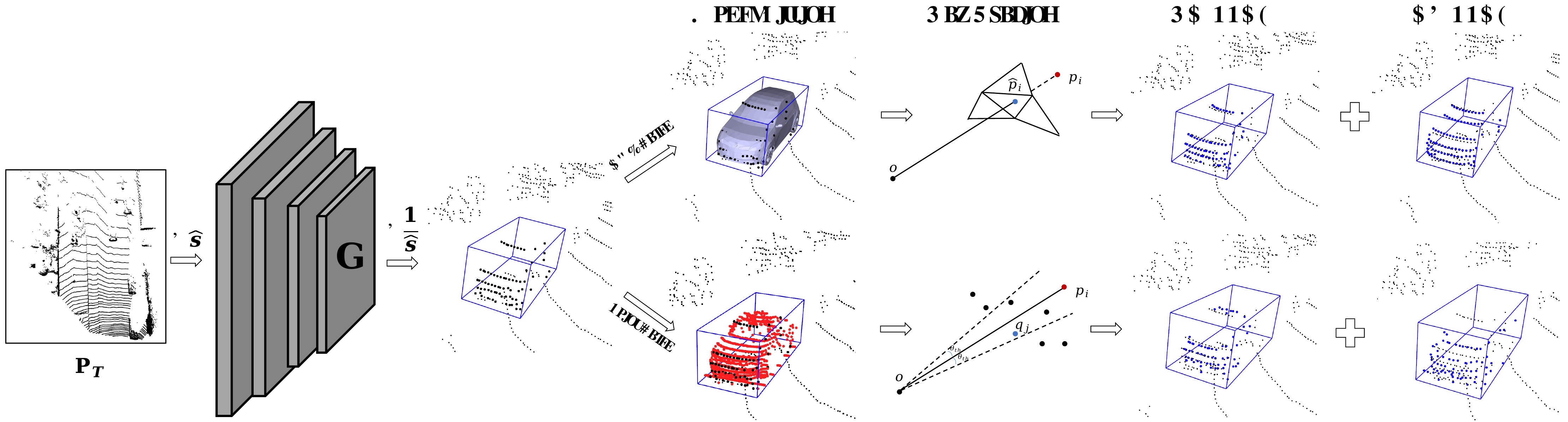}
    	\end{tabular} 
    	\caption{Illustration of our pseudo point clouds generation pipeline. For each target bounding box after PTSN, we first search the 3D model library for the best-fitted 3D model (CAD or points). Then this 3D model is aligned with the bounding box, and 3D ray tracing is applied to generate ray-constrained and constrain-free pseudo point clouds. Black and blue points denote the raw and augmented point clouds, respectively. CAD and points 3D models are represented by the gray meshes and red points, respectively. }
    	\label{fig:demo}
    \end{figure*} 
    
\textbf{Ray-constrained Pseudo Point Clouds Generation.}
Given a pseudo bounding box $\hat{\textbf{b}}$ and the set of original points $\textbf{p}$ inside $\hat{\textbf{b}}$ from the target domain, ray-constrained pseudo point clouds generation (RC-PPCG) tries to generate pseudo point clouds on the same 3D rays as the raw point clouds in order to reduce the deviation between them. More specifically, we adopt the following two steps to perform RC-PPCG.

Firstly, the best-fitted 3D model is selected from the 3D model library $\mathcal{M}$. For CAD-based PPCG, the best-fitted 3D model is the one with minimal size deviation from the pseudo bounding box $\hat{\textbf{b}}$. For point-based PPCG, the best-fitted 3D model is the one with a similar observation angle and minimal distance between point clouds.

Next, the best-fitted 3D model is \textit{transformed} and \textit{scaled} to fit $\hat{\textbf{b}}$ according to the bounding box sizes. The location of the LiDAR sensor in the original point cloud coordinate system is obtained from the sensor library $\mathcal{L}$. Each point $p_i$ in $\textbf{p}$ is connected to the LiDAR sensor to form a 3D ray, and the simulated point $\hat{p}_i$ of $p_i$ is obtained according to the type of the 3D model. If the 3D model is CAD, $\hat{p}_i$ is the intersection point of this 3D ray with the surface of the best-fitted CAD model. If the 3D model is a point cloud, $\hat{p}_i$ is selected from the point cloud as the one with the smallest depth among those points with a deviation angle smaller than a threshold (e.g. $\angle\hat{p}_iop_i<\theta_{th}$ in \fig~\ref{fig:demo}). $\theta_{th}$ is set as twice the average point angle ($360^{\circ}/$Points Per Beam) of the LiDAR sensor. 

The set of simulated points $\hat{\textbf{p}}_{\text{RC}}=\{ \hat{p}_1, \hat{p}_2, ..., \hat{p}_n\}$ is the RC-PPCG pseudo point clouds for the bounding box $\hat{\textbf{b}}$. Note that the original points in each pseudo bounding box will be replaced by the corresponding RC-PPCG pseudo point clouds such that the point clouds and pseudo bounding boxes are consistent for training. The number, density, and distribution of the RC-PPCG pseudo point clouds are quite similar to those of the original points. Therefore, training the network on $(\hat{\textbf{P}}_{\text{RC}}, \hat{\textbf{B}})$ will efficiently improve the performance on the original target samples. 
In practice, we only perform RC-PPCG for pseudo labels whose inside point clouds are less than a threshold, e.g., 300, because the instance-level noise is not likely to exist on those pseudo labels with dense point clouds.

\textbf{Constraint-free Pseudo Point Clouds Generation.} Given a 3D model and a virtual LiDAR sensor, we can also generate pseudo point clouds without the ray constraints. The goal of constraint-free pseudo point clouds generation (CF-PPCG) is to improve the network's performance in challenging scenarios, e.g., far-range objects with very sparse point clouds, by simulating objects that are hard to detect. The first step of CF-PPCG is the same as RC-PPCG, which is to determine the best-fitted model in the model library $\mathcal{M}$. Then, given the field of view, number of beams, and points per beam of the LiDAR sensor, a set of virtual 3D rays are simulated to scan the 3D model. For each 3D ray, the same strategies used in RC-PPCG are applied to generate pseudo point clouds in both CAD-based and point-based cases. The collection of those intersection points is denoted as the CF-PPCG pseudo point clouds $\hat{\textbf{p}}_{\text{CF}}$ for the bounding box $\hat{\textbf{b}}$. Considering that close-range objects are not likely to have noises on the pseudo labels, for each pseudo bounding box $\hat{\textbf{b}} = \{x, y, z, l, w, h, head \}$, we move it to a far-range location $\hat{\textbf{b}}^{'} =\{s\times x, s\times y, z, l, w, h, head\}$ along its direction on the XY plane.
Then, CF-PPCG pseudo point clouds are generated for the bounding box $\hat{\textbf{b}}^{'}$ in the new location. Those points are transformed back to the original location by dividing the X and Y of the pseudo point clouds by $s$. Finally, these transformed points are used as the CF-PPCG pseudo point clouds for $\hat{\textbf{b}}$. Namely, $\hat{\textbf{p}}_{\text{CF}}(\hat{\textbf{b}}) = \hat{\textbf{p}}_{\text{CF}}(s\times \hat{\textbf{b}}) / s$.
Similarly, $(\hat{\textbf{P}}_{\text{CF}}, \hat{\textbf{B}})$ are also utilized to train the network.

In practice, both PTSN and PPCG can be conducted multiple times to improve the performance iteratively as shown in \alg~\ref{alg:dali}. Note that the ray-constrained pseudo point clouds $\hat{\textbf{P}}_{\text{RC}}$ are very similar to the raw point clouds, but contain less noise. Therefore, the raw point clouds $\textbf{P}_{\text{T}}$ in target domain are not used in \alg~\ref{alg:dali} to train the network.
    \begin{algorithm}
    \caption{DALI: Domain Adaptive LiDAR Object Detection via Distribution-level and Instance-level Pseudo Label Denoising}\label{alg:dali}
    \begin{algorithmic}
        \Require a labeled source domain $(\textbf{P}_{\text{S}}, \textbf{B}_{\text{S}})$, an unlabeled target domain
        \Ensure optimized network \textbf{G}, best post-training scale $\hat{s}$ for target domain
        \State \textbf{G} $\gets$ Pre-trained with $(\textbf{P}_{\text{S}}, \textbf{B}_{\text{S}})$
        \For{each iteration}
            \State $\hat{s}$ $\gets$ Post-training size normalization given \textbf{G}
            \State $\hat{\textbf{B}}$ $\gets$ Bounding box prediction given \textbf{G} and $\hat{s}$
            \State $(\hat{\textbf{P}}_{\text{RC}}, \hat{\textbf{B}})$ $\gets$ RC-PPCG
            \State $(\hat{\textbf{P}}_{\text{CF}}, \hat{\textbf{B}})$ $\gets$ CF-PPCG
            \State \textbf{G} $\gets$ Trained with $(\textbf{P}_{\text{S}}, \textbf{B}_{\text{S}})$, $(\hat{\textbf{P}}_{\text{RC}}, \hat{\textbf{B}})$, and $(\hat{\textbf{P}}_{\text{CF}}, \hat{\textbf{B}})$
        \EndFor
    \end{algorithmic}
    \end{algorithm}

\section{Experiments}
Our framework for Domain Adaptive LIDAR (DALI) object detection via distribution-level and instance-level pseudo label denoising is referred to as DALI for the remainder of this section. We use DALI(Point) and DALI(CAD) to represent the variant of DALI with point-based and CAD-based pseudo point clouds generation, respectively.

\subsection{Experimental Setup}
\label{ExperimentalSetup}
    \begin{figure*}
    	\centering
    	\footnotesize
            \begin{tabular}{c}
    		\includegraphics[height=0.18\linewidth]{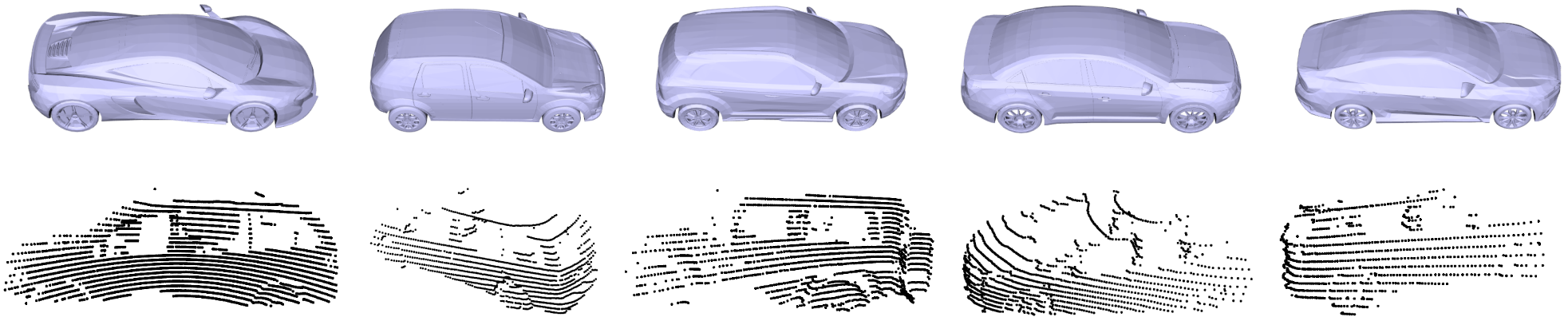} \\
    	\end{tabular} 
    	\caption{Some examples of the CAD-based 3D model and point-based 3D model used in our method. }
    	\label{fig:dataset_info}
    \end{figure*}

\G{
\textbf{Datasets.} In our experiments, we utilized three popular datasets: KITTI~\cite{geiger2012we}, Waymo~\cite{sun2020scalability}, and nuScenes~\cite{caesar2020nuScenes}. Given that (a) LiDAR point cloud datasets provide a sufficient number of data points for objects belonging to the car class; and (b) the car class is the primary focus of many other LiDAR-based object detection methods, we focused our experiments on the car class as well, and we performed domain adaptation on three representative tasks: Waymo $\rightarrow$ KITTI, Waymo $\rightarrow$ nuScenes, and nuScenes $\rightarrow$ KITTI. To reduce training time for the large-scale Waymo dataset, we selected a subset of 1/5 samples for experiments. For CAD-based pseudo point clouds generation, we downloaded the CADillac car dataset~\cite{cadillac} and removed emergency vehicles, old cars manufactured before the year 2000, and uncommon cars with lengths longer than 8m or shorter than 2m. This resulted in a clean dataset with 449 CAD models. For point-based pseudo point clouds generation, we selected annotations with 300 or more internal points as 3D models, resulting in 46,591 and 12,554 3D models for Waymo and nuScenes, respectively. In \fig~\ref{fig:dataset_info}, we show some examples of CAD 3D models and point 3D models. Both types of 3D models have detailed 3D structures, making the pseudo point clouds generated from them reliable for training.
}

\G{
\textbf{Comparison Methods.} We compare our DALI framework with five methods, including 1) Source only, which trains the network with source domain data only and then tested on the target domain dataset; 2) SN~\cite{wang2020train}, which adopts statistical normalization to handle the object-level divergence ; 3) ST3D~\cite{yang2021st3d}, which proposes methods to address both object-level and sensor-level divergences; 4) ST3D++~\cite{yang2022st3d++}, which is an advanced version of ST3D; 5) DTS~\cite{hu2023density}, which is one of the most recent papers published in CVPR 2023.
}

\G{
\textbf{Evaluation Metric.} We adopt the same metric used in ST3D~\cite{yang2021st3d} for evaluation. Namely, we use the official KITTI evaluation metric and report the average precision (AP) over 40 recall positions. The IoU thresholds are 0.7 for both the bird’s eye view (BEV) IoUs and 3D IoUs.
}

\G{
\textbf{Implementation Details.} For the purpose of comparison, we validate our method on the same backbone 3D detectors (SECOND-IoU~\cite{yang2021st3d}) as ST3D based on the code of OpenPCDet~\cite{openpcdet2020}. The Adam~\cite{kingma2014adam} with learning rate $1.5\times 10^{-3}$ and one cycle scheduler is utilized for optimization. The detection range is [-75.2m, 75.2m] for X and Y, and [-2m, 4m] for Z. Also, the input point clouds are shifted in the Z axis such that the origin of the coordinate system is on the ground. The voxel size of SECOND-IoU is (0.1m, 0.1m, 0.15m) for (X, Y, Z) on all the datasets. Considering that it is time consuming to scan the high quality CAD models with a virtual LiDAR, we implement the CUDA version of Möller-Trumbore intersection algorithm~\cite{moller2005fast} to speed up the pseudo point clouds generation procedure. All experiments were conducted on one NVIDIA RTX 3090 GPU. \tab~\ref{tab:time} shows the per-frame time consumption of our PPCG on two target datasets KITTI and nuScenes. We can see that KITTI takes more time than nuScenes due to denser objects on each frame. For the nuScenes→KITTI task, we added the commonly used self-training~\cite{yao2022enhancing,peng2023cl3d,seibold2022reference} at the end of the pipeline, namely, the network is further trained with the target domain and the pseudo labels to better adapt to the target domain.
}

\begin{table}[!ht]
        \centering
    	\caption{Time consumption per frame of our PPCG methods on different datasets including KITTI and nuScenes.}
    	\label{tab:time}
    	\setlength{\tabcolsep}{5.0mm}
    	\begin{tabular}{l c c}
    	\hline
    	Dataset &RC-PPCG &CF-PPCG \\
    	\hline
            KITTI    &0.40s &1.10s\\
            nuScenes &0.07s &0.27s\\
    	\hline
        \end{tabular}
    \end{table}

    \begin{table*}[!ht]
        \centering
    	\caption{LiDAR dataset overview according to the information retrieved from \cite{yang2021st3d}. Note that we use version 1.0 of Waymo Open Dataset and sample only 1/5 frames for training.}
    	\label{tab:datasets_overview}
    	\small
    	\begin{tabular}{c| c| c| c| c| c| c}
    	\hline
    	Dataset &$\#$Beam Ways &Beam Angles &$\#$ Points Per Beam &$\#$ Training Frames &$\#$ Validation Frames &Mean Car Dimensions\\
    	\hline
    	KITTI    &64-beam &[-23.6$^{\circ}$, 3.2$^{\circ}$] &1,843  &3,712   &3,769  &[3.89,1.62,1.53] \\    	
    	\hline
    	Waymo    &64-beam &[-18.0$^{\circ}$, 2.0$^{\circ}$] &2,500  &31,593 &39,987 &[4.66,2.08,1.73] \\
    	\hline
    	nuScenes &32-beam &[-30.0$^{\circ}$, 10.0$^{\circ}$] &781   &28,130 &6,019   &[4.63,1.96,1.73] \\
    	\hline
        \end{tabular}
        \vspace{-0.3cm}
    \end{table*}

\subsection{Comparison on Target Domain}
    \begin{table}[!ht]
        \centering
    	\caption{ Results of different adaptation tasks with SECOND-IoU as the backbone. We report $AP_{BEV}$ and $AP_{3D}$ of the car category at IoU = 0.7. The best adaptation results are indicated in bold. ROS-based and SN-based means the size normalization is performed based on random object scaling (ROS)~\cite{yang2021st3d} and statistical normalization (SN)~\cite{wang2020train} respectively. W, K, n denote Waymo, KITTI, and nuScenes respectively. $\ddagger$ indicates we apply self-training~\cite{yao2022enhancing,peng2023cl3d,seibold2022reference} as an extra procedure at the end of the pipeline.}
    	\label{tab:comparison}
    	\small
    	\setlength{\tabcolsep}{1.5mm}
    	\begin{tabular}{c| c| l| c}
    	\hline
    	Task &Category &Method &$AP_{BEV}/AP_{3D}$ (Target)\\
    	\hline
    	\multirow{11}{*}{\makecell{W → K}}   
            &- & Source Only                           &67.64 / 27.48 \\
            \cline{2-4}
            &\multirow{5}{*}{\makecell{ROS-based}}    	
    	& ST3D~\cite{yang2021st3d}               &82.19 / 61.83 \\
    	&& ST3D++~\cite{yang2022st3d++}          &80.78 / 65.64 \\
            && DTS~\cite{hu2023density}              &\textbf{85.80} / 71.50 \\
            && DALI(Point)                           &85.10 / 74.37 \\
            && DALI(CAD)                             &85.43 / \textbf{74.95} \\
            \cline{2-4}
            &\multirow{5}{*}{\makecell{SN-based}}
            & SN~\cite{wang2020train}                &78.96 / 59.20 \\
    	&& ST3D~\cite{yang2021st3d}              &85.83 / 73.37 \\
    	&& ST3D++~\cite{yang2022st3d++}          &\textbf{86.47} / 74.61  \\
            && DALI(Point)                           &85.46 / 75.10 \\
            && DALI(CAD)                             &85.53 / \textbf{75.32} \\
    	\hline
            \hline
    	
    	\multirow{11}{*}{\makecell{n → K}}  
    	&- & Source Only                           &51.84 / 17.92 \\
            \cline{2-4}
            &\multirow{5}{*}{\makecell{ROS-based}}
    	& ST3D~\cite{yang2021st3d}               &75.94 / 54.13 \\
    	&& ST3D++\cite{yang2022st3d++}           &80.52 / 62.37 \\
            && DTS~\cite{hu2023density}              &81.40 / 66.60 \\
    	&& DALI(Point)$^\ddagger$                           &82.01 / 68.53 \\
    	&& DALI(CAD)$^\ddagger$                             &\textbf{82.10} / \textbf{68.98} \\
            \cline{2-4}
            &\multirow{5}{*}{\makecell{SN-based}}
    	& SN~\cite{wang2020train}                &40.03 / 21.23 \\
    	&& ST3D~\cite{yang2021st3d}               &79.02 / 62.55 \\
    	&& ST3D++~\cite{yang2022st3d++}           &78.87 / 65.56 \\    	
    	&& DALI(Point)$^\ddagger$                           &83.26 / 68.84 \\
    	&& DALI(CAD)$^\ddagger$                             &\textbf{83.43} / \textbf{69.09} \\    
    	\hline
            \hline
    	
    	\multirow{11}{*}{\makecell{W → n}}    
    	&- & Source Only                           &32.91 / 17.24  \\
            \cline{2-4}
            &\multirow{5}{*}{\makecell{ROS-based}}
    	& ST3D~\cite{yang2021st3d}                 &35.92 / 20.19  \\
    	&& ST3D++~\cite{yang2022st3d++}            &35.73 / 20.90  \\
            && DTS~\cite{hu2023density}                &\textbf{41.20} / 23.00 \\
    	&& DALI(Point)                             &35.65 / 22.36  \\
    	&& DALI(CAD)                               &36.81 / \textbf{23.77}  \\        	
            \cline{2-4}
            &\multirow{5}{*}{\makecell{SN-based}}
            & SN~\cite{wang2020train}                &33.23 / 18.57  \\
    	&& ST3D~\cite{yang2021st3d}              &35.89 / 20.38  \\
    	&& ST3D++~\cite{yang2022st3d++}          &36.65 / 22.01  \\
    	&& DALI(Point)                           &35.96 / 22.80  \\
    	&& DALI(CAD)                             &\textbf{36.91} / \textbf{23.53}  \\        
    	\hline
        \end{tabular}
    \end{table}

    \begin{table}[!ht]
        \centering
    	\caption{Performances of different domain adaptation methods on the source domain in the nuScenes → KITTI task with SECOND-IoU as the backbone. We report $AP_{BEV}$ and $AP_{3D}$ of the car category at IoU = 0.7.}
    	\label{tab:comparison_sourcedomain}
    	\small
    	\setlength{\tabcolsep}{1.5mm}
    	\begin{tabular}{c| c| l| c}
    	\hline
    	Task &Category &Method &$AP_{BEV}/AP_{3D}$ (Source)\\
    	\hline
    	\multirow{10}{*}{\makecell{n → K}}  
    	&- & Source Only                           &43.14 / 29.33 \\
            \cline{2-4}
            &\multirow{4}{*}{\makecell{ROS-based}}
    	& ST3D~\cite{yang2021st3d}               &23.39 / 8.42 \\
    	&& ST3D++\cite{yang2022st3d++}           &31.83 / 16.12 \\
    	&& DALI(Point)                           &38.97 / 27.96 \\
    	&& DALI(CAD)                             &\textbf{39.27} / \textbf{28.03}\\
     
            \cline{2-4}
            &\multirow{5}{*}{\makecell{SN-based}}
    	& SN~\cite{wang2020train}                &26.07 / 12.59 \\
    	&& ST3D~\cite{yang2021st3d}              &22.10 / 8.47 \\
    	&& ST3D++~\cite{yang2022st3d++}          &25.42 / 13.52 \\    	
    	&& DALI(Point)                           &39.43 / \textbf{28.21} \\
    	&& DALI(CAD)                             &\textbf{39.66} / 28.18 \\ 
    	\hline
        \end{tabular}
    \end{table}
    
\textbf{Waymo → KITTI.} The domain adaptation task from Waymo to KITTI is relatively easier than other tasks since both datasets have dense and accurate point clouds collected by LiDAR sensors with 64-beams, resulting in smaller sensor-level divergence. The results in \tab~\ref{tab:comparison} show that our DALI(CAD) outperforms other methods in both ROS-based and SN-based categories. Specifically, the AP$_{3D}$ of DALI(CAD) is 0.71 higher than ST3D++ in the SN-based category and 9.31 higher when they are based on ROS. Additionally, DALI(CAD) performs slightly better than DALI(Point) in both categories, mainly due to the more realistic pseudo point clouds generated by CAD-based 3D models compared to point-based models. The performances of both DALI(CAD) and DALI(Point) are similar in the ROS-based and SN-based categories since the ground truth mean object sizes calculated from ROS and SN are close to each other. It is worth noting that our method's significant improvement over ST3D in the ROS category indicates that our post-training size normalization (PTSN) method can better utilize information than the original ROS. The improvement of our DALI method over ST3D in the SN-based category further demonstrates the effectiveness of our method. Our proposed PTSN and PPCG methods are relatively straightforward and easy to implement, providing an additional advantage to our method.

\textbf{nuScenes → KITTI.} The domain adaptation task nuScenes → KITTI is particularly challenging due to significant domain divergences. Specifically, nuScenes employs a 32-beam LiDAR sensor, while KITTI uses a 64-beam LiDAR sensor. Additionally, car sizes in nuScenes are generally larger than those in KITTI. Therefore, we applied the widely used self-training~\cite{yao2022enhancing,peng2023cl3d,seibold2022reference} as an extra procedure at the end of the pipeline to make the network learn more from the target domain. We can see from \tab~\ref{tab:comparison} that our DALI(CAD) method still achieves the best performance in both ROS-based and SN-based categories. Specifically, the AP$_{3D}$ of DALI(CAD) is 3.53 higher than ST3D++ in the SN-based category and 6.61 higher when they are based on ROS.

\textbf{Waymo → nuScenes.} The Waymo → nuScenes domain adaptation task is considered the most challenging one due to the significant divergence between the point clouds in the two datasets. The point clouds in the nuScenes dataset are sparse and inaccurate compared to those in Waymo and KITTI, which makes it difficult to transfer the knowledge learned from the ``easy" samples in Waymo to the ``hard" samples in nuScenes. As shown in \tab~\ref{tab:comparison}, the average performance in AP$_{3D}$ for this task is around 20.0, which is much lower than that for other tasks. However, our DALI methods still achieve the best performance in this task.

\subsection{Cross-domain Performance}
Another advantage of our DALI over previous methods such as SN, ST3D, and ST3D++ is its ability to perform effectively across domains, encompassing both the source and target domains. As discussed in Section~\ref{distribution_level_denoising}, conventional methods typically adjust annotations in the source domain to align with the object sizes in the target domain. However, networks trained with this approach often struggle to excel in the source domain. In contrast, our method, as depicted in \alg~\ref{alg:dali}, trains the network using raw samples from the source domain $(\textbf{P}_{\text{S}}, \textbf{B}_{\text{S}})$ and pseudo samples from the target domain $(\hat{\textbf{P}}_{\text{RC}}, \hat{\textbf{B}})$ and $(\hat{\textbf{P}}_{\text{CF}}, \hat{\textbf{B}})$. This strategy enables competitive performance in both source and target domains. \tab~\ref{tab:comparison_sourcedomain} showcases the performances of various domain adaptation methods on the source domain in the nuScenes → KITTI task. It is evident that our DALI surpasses SN, ST3D, and ST3D++ by a significant margin in the source domain, underscoring the cross-domain capability of our approach for 3D object detection.

\subsection{Ablation Study}    
\begin{figure}
    	\centering
    	\footnotesize
    	\begin{tabular}{cc}
    		 \includegraphics[height=0.40\linewidth]{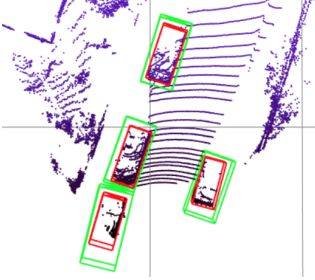} 
    		&\includegraphics[height=0.40\linewidth]{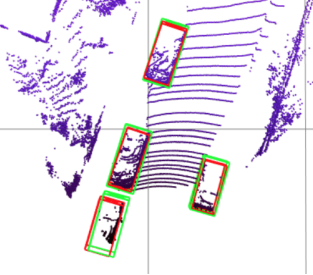} \\
    		(a) Source only  &(b) PTSN(w/ SN) (Our)\\
    	\end{tabular} 
    	\caption{Comparison between Source only and our PTSN(w/ SN) on the domain adaptation task nuScenes → KITTI with backbone SECOND-IoU. The ground truth and predicted bounding boxes are marked in red and green, respectively. It should be clear that our approach provides much more accurate and tighter boxes.}
    	\label{fig:psn_1}
    \end{figure}

\textbf{Post-training Size Normalization.}
To demonstrate the effect of the proposed post-training size normalization (PTSN) method, we compare it with Source only and SN~\cite{wang2020train} on all three domain adaptation tasks with SECOND-IoU as the backbone. From \tab~\ref{tab:psn_table}, we can see that our PTSN(w/ SN) shows significant improvement over both Source only and SN in two sub-tasks: Waymo → KITTI and nuScenes → KITTI. It also can be observed that nuScenes → KITTI shows a much larger improvement than Waymo → KITTI. An explanation for this phenomenon is that our PTSN(w/ SN) scales the whole point clouds; thus, when the scale is larger than 1.0, it has the effect of sparsifying the KITTI point clouds to make them as sparse as those of nuScenes. Methods like SN~\cite{wang2020train} and ROS~\cite{yang2021st3d} that modify the object only may not have such an effect. In the most challenging task Waymo → nuScenes, our PTSN performs worse than the other two methods. The main reason is that we use only 1/5 training samples in Waymo; consequentially, the network does not learn sufficient knowledge about the ``hard" samples that are common in the nuScenes dataset. Theoretically, our PTSN(w/ SN) should achieve better performance than Source only in any tasks, because the post-training size normalization is performed after training the network with Source only. Despite that PTSN(w/ SN) does not perform well on Waymo → nuScenes, the final performance of our method is still the best among all the methods, which shows the effectiveness of incorporating our PTSN and APP approaches. \fig~\ref{fig:psn_1} further shows the advantage of our proposed PTSN method. From \fig~\ref{fig:psn_1} (a) we can see that the predicted bounding boxes (the green ones) are oversized than the ground truth (the red ones) because the network is trained on the nuScenes dataset which has larger car size than KITTI. In \fig~\ref{fig:psn_1} (b), we can see a clear improvement in the bounding boxes which fit the point clouds well after applying our post-training size normalization.

    \begin{table*}[!ht]
        \centering
    	\caption{Comparison between Source only, SN~\cite{wang2020train}, and our PTSN(w/ SN) on different domain adaptation tasks with SECOND-IoU as the backbone.}
    	\label{tab:psn_table}
    	\small
    	\begin{tabular}{l| c | c | c}
    	\hline
    	\multirow{2}{*}{Methods}  &\multicolumn{3}{c}{$AP_{BEV}/AP_{3D}$} \\
    	\cline{2-4}
    	& W → K &n → K &W → n \\
    	\hline
    	(a) Source only                       &67.64 / 27.48 &51.84 / 17.92 &32.91 / 17.24\\
    	(b) SN~\cite{wang2020train}           &78.96 / 59.20 &40.03 / 21.23 &33.23 / 18.57\\
    	(c) PTSN(w/ SN)                       &81.94 / 63.96 &74.30 / 51.18 &27.75 / 11.88 \\
    	\hline
        \end{tabular}
    \end{table*}

    \begin{table}[!ht]
        \centering
    	\caption{Ablation study of our pseudo point clouds generation strategy on domain adaptation task Waymo → KITTI with SECOND-IoU as the backbone. The accuracies are reported based on the performance of SN-based DALI(CAD). \textbf{PTSN} denotes the post-training size normalization with SN. \textbf{RC-PPCG} and \textbf{CF-PPCG} represent the ray-constrained and constraint-free pseudo point clouds generation, respectively. \textbf{PTSN + PPCG (RC\& CF) ($\mathbf{\times}$ 2)} means we perform PTSN and PPCG (RC\& CF) for two iterations. }
    	\label{tab:app_table}
    	\small
    	\setlength{\tabcolsep}{1.3mm}
            \renewcommand*{\arraystretch}{1.2}
    	\begin{tabular}{l| c}
    	\hline
    	Methods  &$AP_{BEV}/AP_{3D}$ \\
    	\hline
    	(a) PTSN only                         &81.94 / 63.96 \\
    	(b) PTSN + RC-PPCG                     &84.38 / 72.06 \\
    	(c) PTSN + CF-PPCG                    &82.71 / 68.89 \\
    	(d) PTSN + PPCG (RC\&CF)               &85.24 / 73.52 \\
    	(e) PTSN + PPCG (RC\&CF) \ $(\times 2)$ &85.53 / 75.32 \\
    	\hline
        \end{tabular}
    \end{table}

    \begin{table}[!ht]
        \centering
    	\caption{Performance of ST3D~\cite{yang2021st3d} and ST3D(w/ SN)~\cite{yang2021st3d} fine-tuned with target domain samples generated by our pseudo point clouds generation (PPCG) method on Waymo → KITTI with SECOND-IoU as the backbone.}
    	\label{tab:finetune_table}
    	\small
    	\begin{tabular}{l| c}
    	\hline
    	Methods  &$AP_{BEV}/AP_{3D}$\\
    	\hline
    	ST3D~\cite{yang2021st3d}                  &82.19 / 61.83 \\
    	ST3D~\cite{yang2021st3d} + PPCG            &82.53 / 65.31 \\
    	ST3D(w/ SN)~\cite{yang2021st3d}           &85.83 / 73.37 \\
    	ST3D(w/ SN)~\cite{yang2021st3d} + PPCG     &85.52 / 74.22 \\
    	\hline
        \end{tabular}
    \end{table}        
    
\textbf{Pseudo Point Clouds Generation.}
\tab~\ref{tab:app_table} shows the ablation study of our pseudo point clouds generation (PPCG) on domain adaptation task Waymo → KITTI. We set PTSN only as the baseline and analyze the performances by adding different combinations of RC-PPCG (ray-constrained pseudo point clouds generation) and CF-PPCG (constraint-free pseudo point clouds generation). Comparing (b) and (c) with (a), we can see that both types of pseudo point clouds generation can improve the performance. Generally, RC-PPCG performs better than CF-PPCG, while their combination in (d) outperforms both of them. The reason is that RC-PPCG is designed to be similar to the original target point clouds, while CF-PPCG focuses on hard-predicted objects. Therefore, RC-PPCG will contain more target information than CF-PPCG, which leads to better performance. Scenario (e) in the same table shows the results after two iterations. Namely, as shown in \alg~\ref{alg:dali}, we take the network trained by PTSN + PPCG (RC\&CF) as the pre-trained model, and perform PTSN and PPCG (RC\&CF) again to achieve the results in (e). Comparing (a), (d), and (e), we can see that the increments in AP$_{BEV}$ and AP$_{3D}$ of the second iteration (0.29 / 2.80) are not comparable to those of the first iteration (3.59 / 11.36), but are still significant enough. In practice, we find that conducting PTSN and PPCG (RC\&CF) for two iterations achieves a good balance between time consumption and accuracy. Therefore, we use this setting for all our experiments in \tab~\ref{tab:comparison}.

\textbf{PPCG as Post-Processing.}
Given a pre-trained network, our proposed PPCG can be adopted as a post-processing method to further improve the performance. To demonstrate this, we download the pre-trained models of ST3D~\cite{yang2021st3d} and ST3D(w/ SN)~\cite{yang2021st3d} on Waymo → KITTI and freeze the backbone parts in the network to fine tune the head parts with the pseudo target samples generated by our CAD-based PPCG method. \tab~\ref{tab:finetune_table} shows the results of ST3D~\cite{yang2021st3d} and ST3D(w/ SN)~\cite{yang2021st3d}, from which we can see that simply adding PPCG to existing methods to fine tune the head parts of the network can significantly improve the performance. The rationale is that the pseudo point clouds and pseudo bounding boxes in our PPCG samples are consistent with each other. Therefore, they can provide useful information to guide the head parts to learn a correct mapping from the feature space to the label space.

\textbf{Robustness to Backbone.}
To demonstrate the robustness of our method against the variation of architecture, we test our method on the task nuScenes → KITTI with PV-RCNN~\cite{shi2020pv} as the backbone. As can be seen from \tab~\ref{tab:pvrcnn}, our DALI(CAD) achieves the best performance in AP$_{3D}$, which is 2.49 higher than ST3D~\cite{yang2021st3d} in the SN-based category and 4.22 higher in the ROS-based category. This proves the robustness of our method.

    \begin{table}[!ht]
        \centering
    	\caption{Performance of our method on domain adaptation task nuScenes → KITTI with PV-RCNN~\cite{shi2020pv} as the backbone. The accuracies are reported based on the performance of SN-based DALI(CAD).}
    	\label{tab:pvrcnn}
    	\small
    	\setlength{\tabcolsep}{3mm}
    	\begin{tabular}{c| c | c}
    	\hline
    	Category &Methods  &$AP_{BEV}/AP_{3D}$\\
    	\hline    	
    	- &Source only                               &68.15 / 37.17 \\   
            \hline
            \multirow{2}{*}{\makecell{ROS-based}}
    	&ST3D~\cite{yang2021st3d}                  &78.36 / 70.85 \\
            &DALI(CAD)                                 &83.86 / 75.07 \\
            \hline
            \multirow{3}{*}{\makecell{SN-based}}
            &SN~\cite{wang2020train}                   &60.48 / 49.47 \\
    	&ST3D~\cite{yang2021st3d}                  &84.29 / 72.94 \\
    	&DALI(CAD)                                 &84.12 / 75.43 \\
    	\hline
        \end{tabular}
    \end{table}

    \begin{figure*}
        \begin{tabular}{c}
            \includegraphics[width=1.0\linewidth]{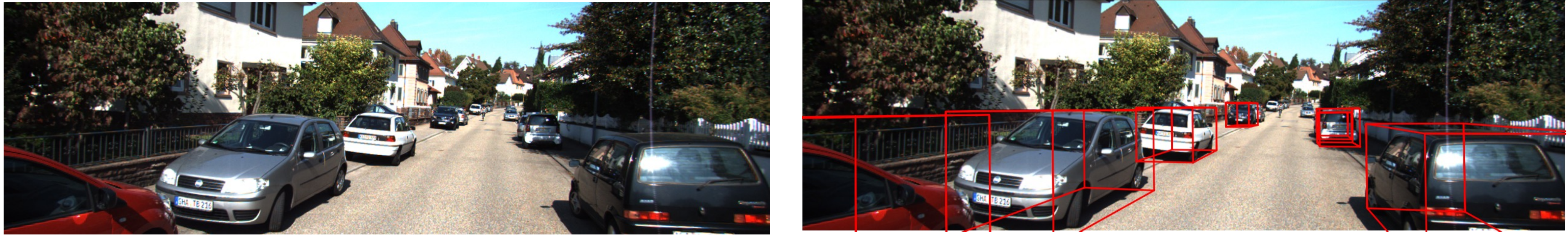} \\
            (a) left: 2D image, right: ground truth 3D bounding boxes drawn on image \\
            \includegraphics[width=1.0\linewidth]{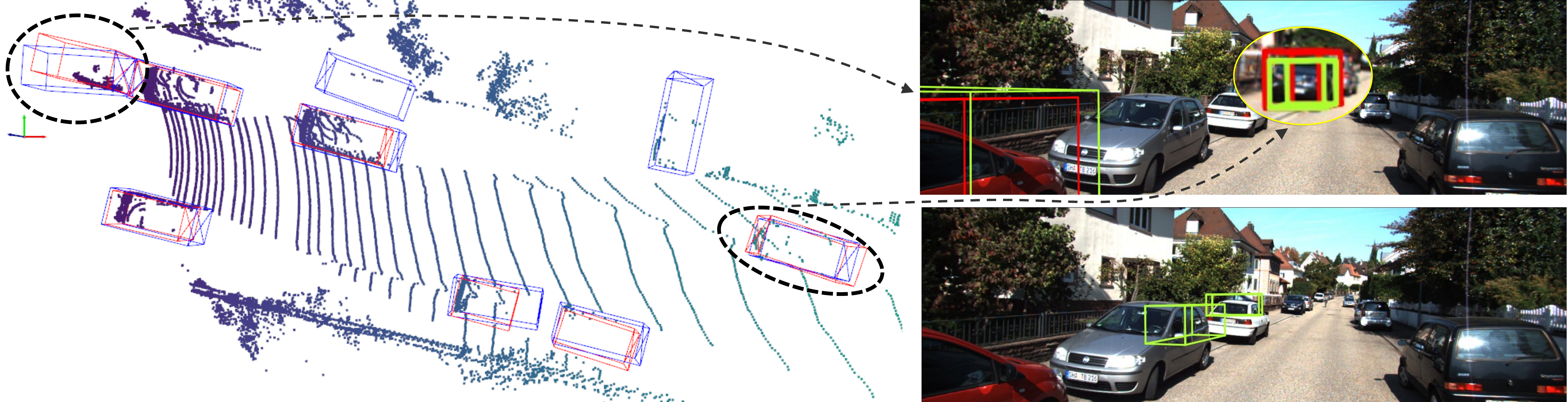}\\ 
            (b) left: 3D results of DALI, upper right: demonstration of the dashed black circle areas on image, bottom right: FPs of \\
            DALI drawn on image\\
            \includegraphics[width=1.0\linewidth]{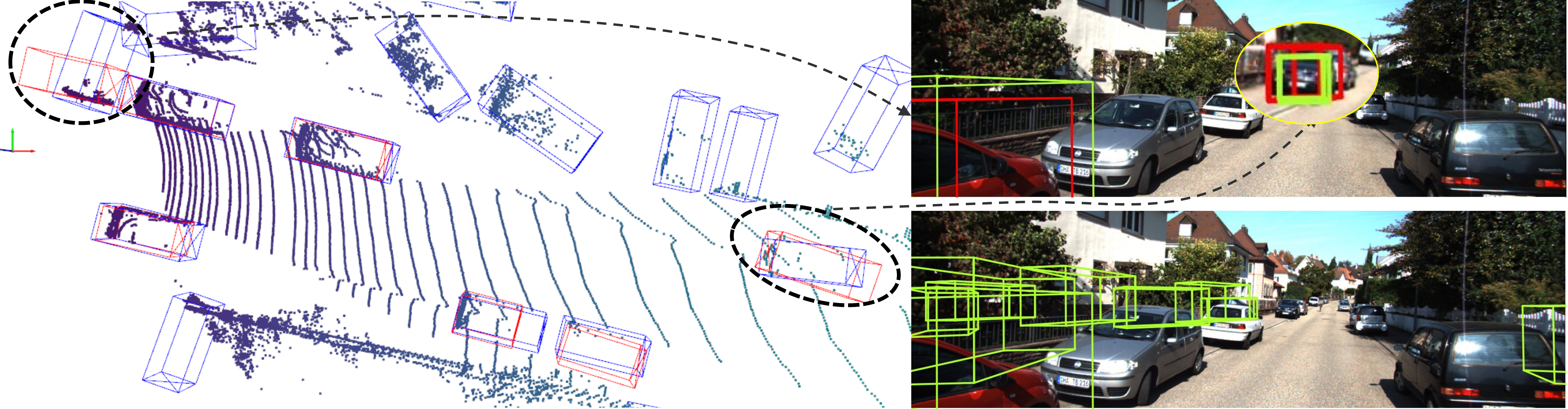}\\ 
            (c) left: 3D results of ST3D++~\cite{yang2022st3d++}, upper right: demonstration of the dashed black circle areas on image, bottom right: FPs of \\
            ST3D++~\cite{yang2022st3d++} drawn on image\\
            \includegraphics[width=1.0\linewidth]{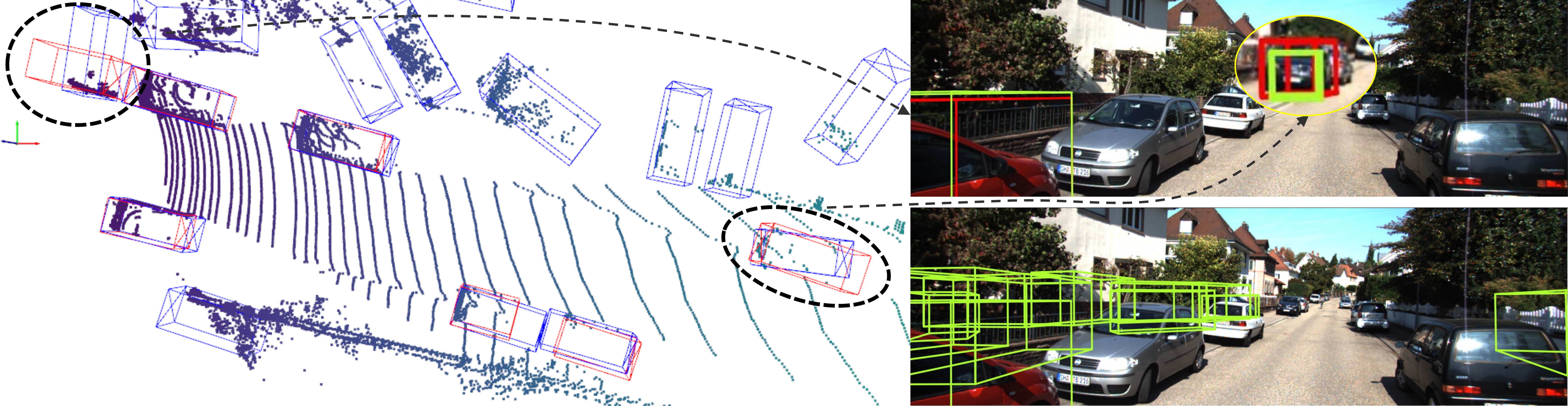}\\ 
            (d) left: 3D results of ST3D~\cite{yang2021st3d}, upper right: demonstration of the dashed black circle areas on image, bottom right: FPs of \\
            ST3D~\cite{yang2021st3d} drawn on image\\
        \end{tabular} 
        \caption{Examples of detection results of DALI, ST3D++~\cite{yang2022st3d++}, and ST3D~\cite{yang2021st3d} on the task nuScenes → KITTI with backbone SECOND-IoU. Those predicted bounding boxes with no overlap with the ground truth are considered false positives and denoted as FPs. Red bounding boxes denote ground truth in both 3D point cloud and 2D image. Predicted bounding boxes are marked with blue and green in 3D point cloud and 2D image respectively. Our DALI framework can generate more true-positives and significantly less false-positives (FPs) in comparison with ST3D++~\cite{yang2022st3d++} and ST3D~\cite{yang2021st3d}.}
        \label{fig:example_results}
    \end{figure*}

\subsection{Visual Results} 
\fig~\ref{fig:example_results} illustrates the detection results of our proposed DALI, ST3D++~\cite{yang2022st3d++}, and ST3D~\cite{yang2021st3d} methods on the nuScenes → KITTI task, using the SECOND-IoU backbone. We utilized the pre-trained models provided by the authors of ST3D++ and ST3D to generate the detection results. The figure shows that all three methods can achieve similar performance when the objects are easily detectable, such as those on the street with dense point clouds. However, for challenging objects such as the closest car on the left side, which is only partially scanned by LiDAR, and the farthest one on the left side with sparse point clouds, both ST3D++ and ST3D fail to generate the correct bounding boxes. Nevertheless, our method can still work well in such challenging conditions because our ray-constrained pseudo point clouds generation can correct partially scanned point clouds to make them consistent with pseudo bounding boxes for network training. Furthermore, our constraint-free pseudo point clouds generation can significantly increase the number of training samples with sparse point clouds, thereby improving the network's performance on distant objects. Additionally, it is worth noting that both ST3D++ and ST3D generate many more false positive detections than our method. However, the evaluation metric of KITTI primarily focuses on the recall rate of the results. Thus, the advantage of our DALI in terms of fewer false positive detections may not be reflected in \tab~\ref{tab:comparison}.

\begin{figure*}
    	\centering
    	\footnotesize
    	\begin{tabular}{c}
    		 \includegraphics[width=0.90\linewidth]{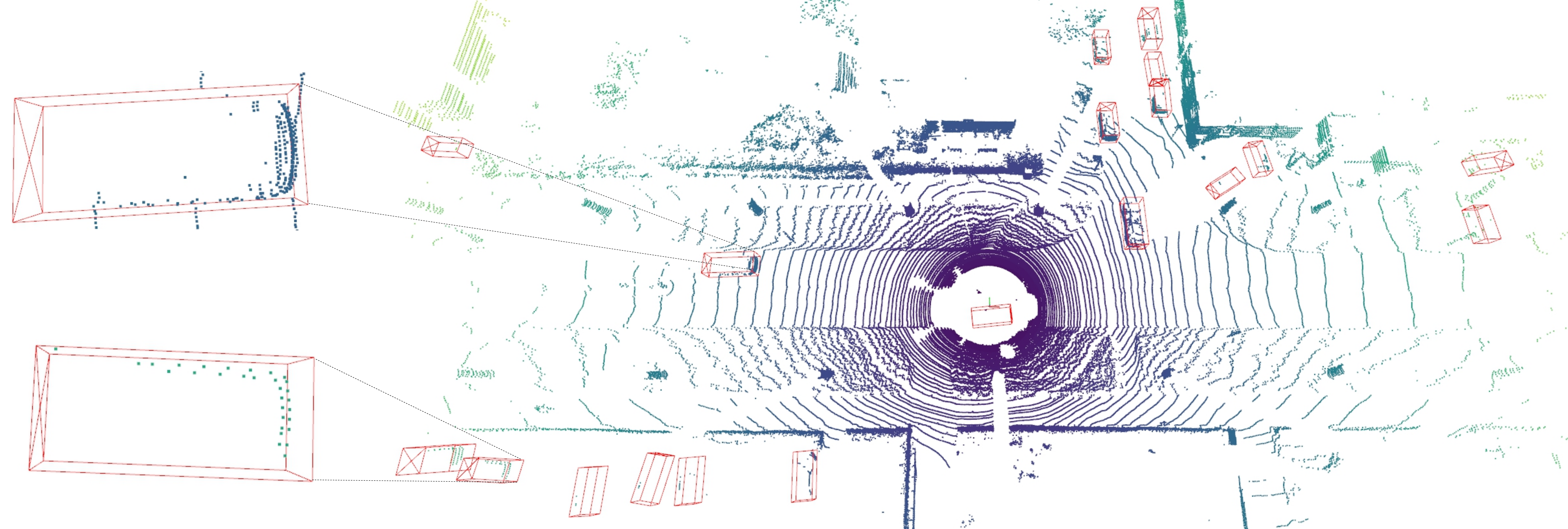} \\
    		 (a) Ray-constrained PPCG\\
    		 \includegraphics[width=0.90\linewidth]{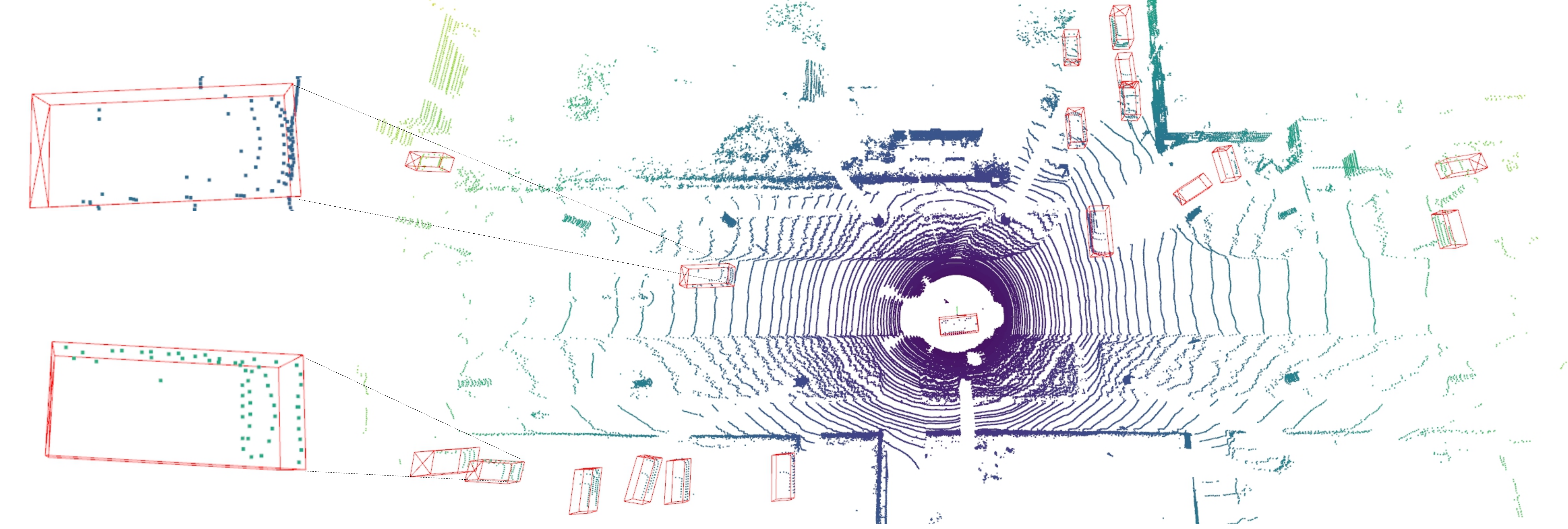} \\
    		 (b) Constraint-free PPCG\\
    	\end{tabular} 
    	\caption{Examples of ray-constrained and constraint-free pseudo point clouds and corresponding pseudo bounding boxes (in red) used in the first iteration on the task nuScenes → KITTI with backbone SECOND-IoU.}
    	\label{fig:app_2}
    \end{figure*}

\fig~\ref{fig:app_2} shows examples of training samples generated by ray-constrained and constraint-free pseudo point clouds generation methods for the task nuScenes → KITTI. We can observe that the ray-constrained PPCG can generate point clouds similar to the original point's distribution, e.g., dense for close-range objects and sparse for distant objects. This is because the simulated 3D rays used to scan the 3D models are generated based on the target domain point clouds. For constraint-free PPCG, we can see that all the generated pseudo point clouds share the same sparsity regardless of their distance to the LiDAR sensor, in this way, the diversity of the training samples is increased and also the network's performance in challenging scenarios will be improved.

\subsection{Discussion and Failure Cases} 
Our pseudo point cloud generation (PPCG) strategies rely on the geometrical information of the target object. Therefore, they are limited by the inherent geometrical characteristics of the target object. In our experiments, we found that PPCG works well for rigid objects like cars and trucks but may struggle with non-rigid objects like pedestrians. This is reasonable because non-rigid objects usually have complex geometrical shapes, making them difficult to simulate in PPCG. Additionally, our method has difficulty simulating the noisy LiDAR point clouds caused by adverse weather conditions like rain and fog, limiting our application in cross-weather domain adaptation tasks. In future work, we plan to explore ways to apply our method to non-rigid objects and cross-weather domain adaptation.

\section{Conclusion}
This paper proposes a novel framework for Domain Adaptive LIdar-based (DALI) 3D object detection that includes two strategies, post-training size normalization (PTSN), to address distribution-level noise, as well as a pseudo point clouds generation (PPCG) method to handle instance-level noise for unsupervised domain adaptation for LiDAR object detection. PTSN is achieved by converting the prediction results on the target domain as a function of scale and selecting the optimal unbiased scale $\hat{s}$ that makes the predicted mean object size equal to the estimated ground truth mean object size. The optimal unbiased scale is then used to generate pseudo bounding boxes for the target domain samples. Given the pseudo bounding boxes, our PPCG method generates both ray-constrained and constraint-free pseudo point clouds that are consistent with the pseudo bounding boxes for training. Comprehensive experiments on several popular benchmarks demonstrate the effectiveness of our proposed method. In general, the proposed method could arguably be limited by the accuracy of the LiDAR sensor and the 3D model. Generally, the higher the accuracy of the LiDAR sensor and 3D model, the better the performance. Fortunately, during our experiments, we found that the performance of the proposed DALI framework is not very sensitive to the LiDAR sensor and 3D model. This means that the performance of our method will still be competitive with the state-of-the-art even with poor LiDAR sensor and 3D model. In future work, we plan to extend the performance evaluation of our method to other objects such as pedestrians and bicycles.

\section{Acknowledgement}

This work was supported in part by grants from the US Department of Transportation and the MSU Research Foundation.
    
{\small
\bibliographystyle{ieeetr}
\bibliography{egbib}
}

\begin{wrapfigure}{l}{25mm} \includegraphics[width=1in]{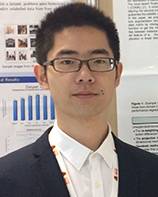}
\end{wrapfigure}\par
\textbf{Xiaohu Lu} received the B.S. degree and M.S. degree in the School of Remote Sensing and Information Engineering, Wuhan University, China, in 2014 and 2017, respectively. He is currently pursuing the Ph.D. degree in Electrical and Computer Engineering, Michigan State University, MI, USA. His research interests include domain adaptation, 3D object detection, 3D point cloud processing, and structure from motion.\par

\begin{wrapfigure}{l}{25mm} \includegraphics[width=1in]{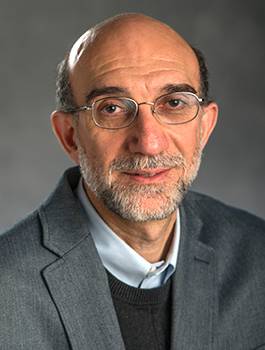}
\end{wrapfigure}\par
\textbf{Hayder Radha} (Fellow, IEEE) received the Ph.M. and Ph.D. degrees from Columbia University, New York, NY, USA, in 1991 and 1993, respectively. He was a Fellow and Principal Member of Research Staff at Philips Research from 1996 to 2000 and a Distinguished Member of Technical Staff and MTS at Bell Laboratories from 1986 to 1996. He is currently a MSU Foundation Professor and the Director of the Connected and Autonomous Networked-Vehicles for Active Safety (CANVAS) Program, Michigan State University. He is a recipient of the Sony Research Award, the Amazon Research Award, the Semiconductor Research Consortium Award, two Google Faculty Research Awards, two Microsoft Research Awards, the AT\&T Bell Labs Ambassador and AT\&T Circle-of-Excellence Awards, and the William J. Beal Outstanding Faculty Award.\par

\end{document}


\title{UDA3D: Unsupervised Domain Adaptation for 3D Object Detection with Posterior Size Normalization and Augmented Pseudo Points \\
Supplementary Material}

\maketitle

In this supplementary material, we provide further information on our proposed posterior size normalization and augmented pseudo points, as well as additional figures and tables to demonstrate the performance of our method.


\section{Posterior Size Normalization}
\fig~\ref{fig:psn_1} shows the results of the method with and without posterior size normalization on the domain adaptation task nuScenes → KITTI with backbone SECOND-IoU. From \fig~\ref{fig:psn_1} (a) we can see that the predicted bounding boxes are oversized than the ground truth because the network is trained on the nuScenes dataset which has larger car size than KITTI. In \fig~\ref{fig:psn_1} (b), we can see clear improvement on the bounding boxes which fit the point clouds well after applying our posterior size normalization. One may wonder what will happen if we scale the bounding boxes in \fig~\ref{fig:psn_1} (a) to make the posterior mean size equal to the prior mean. \tab~\ref{tab:psn_scale} shows the results of bounding box scaling and PSN on task nuScenes → KITTI with backbone SECOND-IoU, from which we can see that simply scaling the bounding boxes is not as effective as PSN which scales the point clouds. The reason is that by scaling the point clouds, PSN can make the objects in the target domain similar to those in the source domain. As a result, the network trained on the source domain will generate better predictions for the scaled target samples.

    \begin{figure*}
    	\centering
    	\footnotesize
    	\begin{tabular}{cc}
    		 \includegraphics[width=0.45\linewidth]{latex/figures_supplementary/Picture8.png} 
    		&\includegraphics[width=0.45\linewidth]{latex/figures_supplementary/Picture9.png} \\
    		(a) Source only  &(b) PSN (Our)\\
    	\end{tabular} 
    	\caption{Comparison between Source only and our PSN on the domain adaptation task nuScenes → KITTI with backbone SECOND-IoU.}
    	\label{fig:psn_1}
    \end{figure*} 
    
    \begin{table}[!ht]
        \centering
    	\caption{Performance of bounding box scaling and our PSN on nuScenes → KITTI with SECOND-IoU as backbone.}
    	\label{tab:psn_scale}
    	\small
    	\begin{tabular}{l| c}
    	\hline
    	Methods  &AP_{BEV}/AP_{3D}\\
    	\hline
    	Source only                             &51.84 / 17.92 \\
    	Bounding box scaling                    &64.65 / 18.22 \\
    	PSN                                     &74.30 / 51.18 \\
    	\hline
        \end{tabular}
    \end{table}  

\section{Augmented Pseudo Points}
\fig~\ref{fig:app_1} shows the 3D point clouds generated by scanning the CAD car model with virtual KITTI and nuScenes LiDARs in different distances. We can observe that the density of point clouds drops when the distance of object to the sensor increases. The augmented points of KITTI are typically denser than those of nuScenes because KITTI has more beams and points per beam. Moreover, we can also find that augmented points are usually dense and complete for these nearby objects, which means that they are rather easy for the network to learn. Hence, for Type II augmentation, we first move the object to a longer-range location to simulate the augmented points and then transform these points back to the original location such that our Type II augmented points can provide more ``hard" smaples for training.

As we all known, the time complexity of scanning a CAD model formed by $m$ triangles with a LiDAR sensor that has $n$ rays is $\mathcal{O}(mn)$, which is really slow in our case where the average number of triangles in the high quality CAD models is 67,735. In order to speed up the process, we develop the CUDA version of the Möller-Trumbore intersection algorithm~\cite{moller2005fast}, and load all the CAD models to the memory to speed up the augmentation. \tab~\ref{tab:app_time} shows the time consumption for Type I and Type II augmentation on KITTI dataset. We can see that the time needed for data augmentation is 86 minutes in total for KITTI, which is really efficient considering that it usually takes dozens of hours to train the network for the domain adaptation task. We can also find that it takes only 0.26 second to generate the Type I augmentation for one frame, while this number is 1.13 for Type II. The reason why Type II takes more time than Type I is that the former uses all the simulated rays to intersect with the CAD triangles while the later connects only the points in each bounding box to the sensor to form the rays. With much more rays needed to be processed, Type II inevitably takes more time.


\fig~\ref{fig:app_2} shows the examples of training samples generated by Type I and Type II augmentation methods for the task nuScenes → KITTI.

    \begin{figure}
    	\centering
    	\footnotesize
    	\begin{tabular}{cc}
    		 \includegraphics[height=0.46\linewidth]{latex/figures_supplementary/Picture1.png} 
    		&\includegraphics[height=0.46\linewidth]{latex/figures_supplementary/Picture2.png} \\
    		(a) KITTI 64 beams at 10m &(b) KITTI 64 beams at 20m \\
    		 \includegraphics[height=0.46\linewidth]{latex/figures_supplementary/Picture3.png} 
    		&\includegraphics[height=0.46\linewidth]{latex/figures_supplementary/Picture4.png} \\
    		(c) nuScenes 32 beams at 10m &(d) nuScenes 32 beams at 20m \\
    	\end{tabular} 
    	\caption{Demonstration of the Type II 3D points generated by scanning the CAD model with different LiDAR sensor at different distances.}
    	\label{fig:app_1}
    \end{figure}


    \begin{figure*}
    	\centering
    	\footnotesize
    	\begin{tabular}{c}
    		 \includegraphics[height=0.40\linewidth]{latex/figures_supplementary/typeI.png} \\
    		 (a) Type I \\
    		 \includegraphics[height=0.40\linewidth]{latex/figures_supplementary/typeII.png} \\
    		 (b) Type II \\
    	\end{tabular} 
    	\caption{Examples of Type I and Type II augmented pseudo points and corresponding bounding boxes used in the first iteration on the task nuScenes → KITTI with backbone SECOND-IoU.}
    	\label{fig:app_2}
    \end{figure*} 
    
    \begin{table}[!ht]
        \centering
    	\caption{Time consumption for Type I and Type II augmentation methods on KITTI datasets with backbone SECOND-IoU in the first iteration.}
    	\label{tab:app_time}
    	\small
    	\setlength{\tabcolsep}{2.0mm}
    	\begin{tabular}{c| c| c}
    	\hline
    	Method &Time total (min) &Time per frame (s)\\
    	\hline
    	Type I      &16  &0.26  \\
    	Type II     &70  &1.13  \\
    	\hline
        \end{tabular}
    \end{table}  
    
\section{Performance on Backbone PVRCNN}
To demonstrate the robustness of our method again the variation of architecture, we test our method on the task nuScenes → KITTI with PV-RCNN~\cite{shi2020pv} as backbone. As can be seen from \tab~\ref{tab:pvrcnn}, our UDA3D achieves the best performance in AP$_{3D}$, which is 2.49 higher than ST3D (w/ SN)~\cite{yang2021st3d} in the second place and 4.58 higher than the original ST3D. As to AP$_{BEV}$, the performance of our UDA3D is close to that of ST3D (w/ SN)~\cite{yang2021st3d} in the first place. The performance of our UDA3D in \tab~\ref{tab:pvrcnn} demonstrates the robustness of our method.

    \begin{table}[!ht]
        \centering
    	\caption{Performance of our method on domain adaptation task nuScenes → KITTI with PV-RCNN~\cite{shi2020pv} as backbone.}
    	\label{tab:pvrcnn}
    	\small
    	\setlength{\tabcolsep}{3mm}
    	\begin{tabular}{l| c}
    	\hline
    	Methods  &AP_{BEV}/AP_{3D}\\
    	\hline
    	
    	Source only                               &68.15 / 37.17 \\
    	SN~\cite{wang2020train}                   &60.48 / 49.47 \\
    	ST3D~\cite{yang2021st3d}                  &78.36 / 70.85 \\
    	ST3D (w/ SN)~\cite{yang2021st3d}          &84.29 / 72.94 \\
    	UDA3D (Our)                               &84.12 / 75.43 \\
    	\hline
        \end{tabular}
        \vspace{-1.0em}
    \end{table}

{\small
\bibliographystyle{ieee_fullname}
\bibliography{egbib}
}